%% file: acl.tex
\pdfoutput=1

\documentclass[11pt]{article}

\usepackage{acl}

\usepackage{times}
\usepackage{makecell}
\usepackage{latexsym}
\usepackage{multirow}
\usepackage{graphicx}
\usepackage{caption}
\usepackage{subcaption}
\usepackage{amssymb, amsmath}
\usepackage[ruled,vlined]{algorithm2e}
\usepackage[shortlabels]{enumitem}
\usepackage{tabularx,ragged2e}

\newcommand{\te}{\textbf{TE-Wiki}}

\usepackage[T1]{fontenc}

\usepackage[utf8]{inputenc}

\usepackage{microtype}

\title{Towards Open-Domain Topic Classification}

\author{Hantian Ding$^1$, Jinrui Yang$^{1,2}$, Yuqian Deng$^1$, Hongming Zhang$^1$, Dan Roth$^1$ \\
  $^1$University of Pennsylvania, 
   $^2$University of Melbourne \\
  \small\texttt{\{hantian2, jinruiy, yuqiand, hzhangal, danroth\}@seas.upenn.edu} \\}

\begin{document}
\maketitle
\begin{abstract}
We introduce an open-domain topic classification system that accepts user-defined taxonomy in real time. Users will be able to classify a text snippet with respect to any candidate labels they want, and get instant response from our web interface. To obtain such flexibility, we build the backend model in a zero-shot way. By training on a new dataset constructed from Wikipedia, our label-aware text classifier can effectively utilize implicit knowledge in the pretrained language model to handle labels it has never seen before. We evaluate our model across four datasets from various domains with different label sets. Experiments show that the model significantly improves over existing zero-shot baselines in open-domain scenarios, and performs competitively with weakly-supervised models trained on in-domain data.\footnote{Interactive online demo at \url{https://cogcomp.seas.upenn.edu/page/demo_view/ZeroShotTC}}\footnote{Code and data available at \url{http://cogcomp.org/page/publication_view/980}}
\end{abstract}

\input{sections/intro.tex}
\input{sections/relevant_work.tex}
\input{sections/methodology.tex}

\input{sections/experiment.tex}

\input{sections/analysis.tex}

\input{sections/crowdsourcing.tex}

\input{sections/conclusion.tex}

\input{sections/ack.tex}

\bibliography{anthology.bib}
\bibliographystyle{acl_natbib}

\newpage
\input{sections/appendix}

\end{document}

%% file: sections/intro.tex
\section{Introduction}

Text classification is a fundamental natural language processing problem, with one of its major applications in topic labeling \cite{Lang95,wang-manning-2012-baselines}.
Over the past decades, supervised classification models have achieved great success in closed-domain tasks with large-scale annotated datasets \cite{DBLP:conf/nips/ZhangZL15, tang-etal-2015-document, yang-etal-2016-hierarchical}. However, they are no longer effective in open-domain scenarios where the taxonomy is unbounded. Retraining the model for every new label set often incurs prohibitively high cost in the sense of both annotation and computation. By contrast, having one classifier that is flexible with unlimited labels can save such tremendous efforts while keeping the solution simple. Therefore, in this work, we build a system for open-domain topic classification that can classify a given text snippet into any categories defined by users.

At the core of our system is a zero-shot text classification model. While supervised models are typically insensitive to class names, a zero-shot model is usually label-aware, meaning that it can understand label semantics directly from the name or definition of the label, without accessing any annotated examples. Our model \te~combines a \textbf{T}extual \textbf{E}ntailment (TE) formulation with Wikipedia finetuning. Specifically, we construct a new dataset that contains three million article-category pairs from Wikipedia's subcategory graph, and finetune a pretrained language model (e.g. BERT) to predict the entailment relations between articles and their associated categories. We simulate the diversity in open-domain classification with the wide coverage of Wikipedia, while preserving label-awareness through an entailment framework. 

In our benchmarking experiments, \te~outperforms all previous zero-shot methods on four benchmarks from different domains. It also shows competitive performance against weakly-supervised models trained on in-domain data. By learning from Wikipedia, our method does not require any data that is specifically collected from the evaluation domains. On the other hand, since our model is label-aware, it can flexibly classify text pieces into any labels outside Wikipedia. 

Finally, we compare our system against humans for further insights. We show that even humans are sometimes confused by ambiguous labels through a crowdsourcing study, which explains the performance gap between open-domain and supervised classification. The gap is reduced significantly when label meanings are clear and well aligned with the semantics of text. We also use an example to illustrate the negative effect of a bad label name. Through the analysis, we demonstrate the importance of choosing proper label names in open-domain classification.

\begin{figure*}[t]
    \centering
    \includegraphics[width=0.8\textwidth]{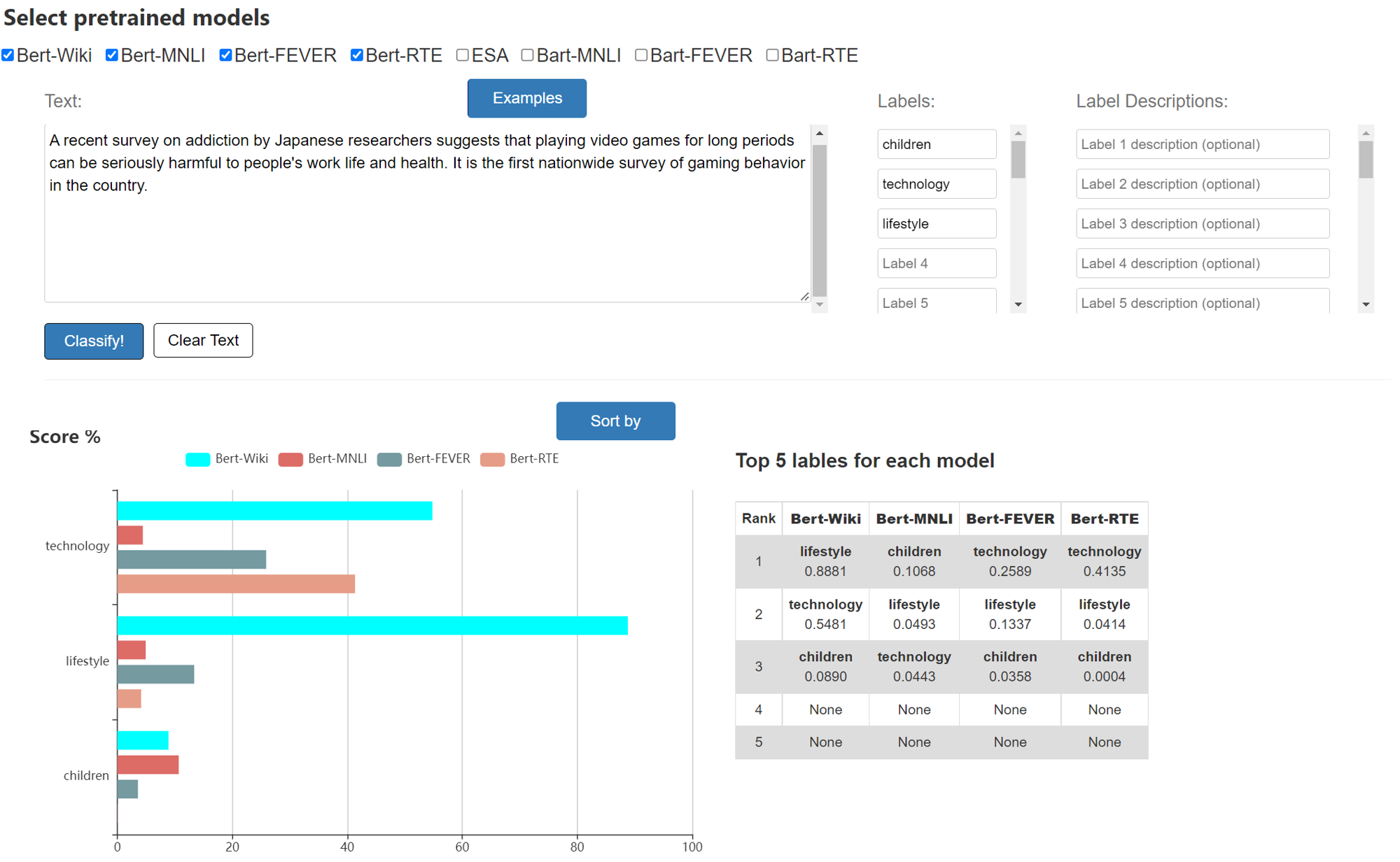}
    \vspace{-0.05in}
    \caption{An overview of our open-domain topic classification system. Users can choose multiple models (\textit{top}), and define their own text input and candidate labels (\textit{middle}). Prediction results from different models are displayed in the bar chart and the table (\textit{bottom}).}
    \vspace{-0.15in}
    \label{fig:demo}
\end{figure*}

%% file: sections/relevant_work.tex
\section{Related Work}

Open-domain zero-shot text classification was first studied in the NLP domain in~\cite{CRRS08} (under the name ``dataless classification") as a method that classifies solely based on a general knowledge source and does not require any in-domain data, whether labeled or not. It was proposed to embed both the text and labels into the same semantic space, via Explicit Semantic Analysis, or ESA \cite{GabrilovichMa07}, and pick the label with the highest relevance score. This idea was further extended to hierarchical \cite{SongRo14} and cross-lingual \cite{SUPR16} text classification. Later on, \cite{YinHaRo19} called this protocol ``label fully unseen" and proposed an entailment approach to transfer knowledge from textual entailment to text classification. It formulates an $n$-class classification problem as $n$ binary entailment problems by converting labels into hypotheses and the text into the premise, and selects the premise-hypothesis pair with highest entailment score. More recently, another concurrent work \cite{chu2021natcat} proposed to explore resources from Wikipedia for zero-shot text classification, but with a different formulation. 

There are many other methods that also require less labeling than supervised classification, though in slightly different settings. For example, previous works have explored to generalize from a set of known classes (with annotation) to unknown classes (without annotation) using word embeddings of label names \cite{DBLP:journals/corr/abs-1712-05972, xia-etal-2018-zero, liu-etal-2019-reconstructing}, class correlation on knowledge graphs \cite{rios-kavuluru-2018-shot, zhang-etal-2019-integrating}, or joint embeddings of documents and labels \cite{DBLP:conf/aaai/NamMF16}. Besides, Weakly supervised approaches \cite{mekala-shang-2020-contextualized, meng-etal-2020-text} learn from an unlabeled, but in-domain training set. Given a set of pre-defined labels, a label-aware knowledge mining step is first applied to find class-specific indicators from the corpus, followed by another self training step to further enhance the model by propagating the knowledge to the whole corpus. However, none of these approaches are suitable for building an open-domain classification system. They either require domain-specific annotation or knowing test labels beforehand. 

%% file: sections/methodology.tex
\section{System Description}

\begin{figure*}[t]
    \centering
    \includegraphics[width=0.8\textwidth]{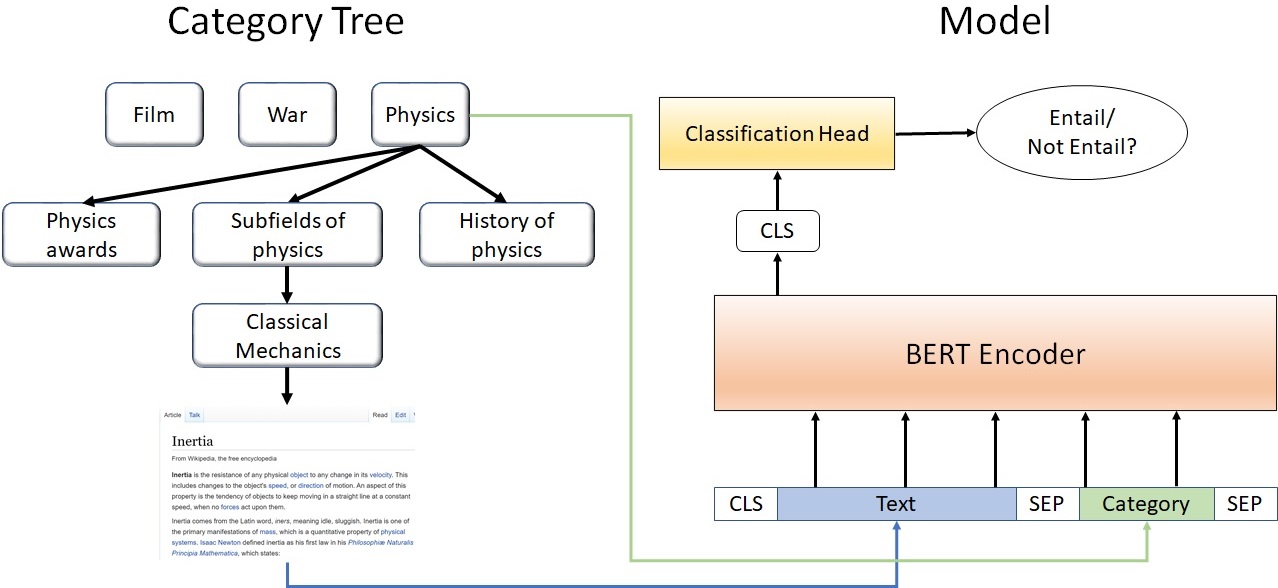}
    \vspace{-0.05in}
    \caption{An overview of our proposed \te. \textit{Left}: the data collection process. For each of the top-level categories, we run DFS to find its descendant categories as well as their member articles. These articles are paired with the root category for model input. \textit{Right}: the model architecture. We use BERT for sequence classification. The article text is concatenated with the category name to feed into a BERT encoder. The classification head takes the output embedding of the "[CLS]" token to classify the input text-category pair.}
    \vspace{-0.15in}
    \label{fig:model}
\end{figure*}

We present details about our open-domain topic classification system, starting with an overview of our web interface, followed by the backend model. 

\subsection{User Interface}

Figure \ref{fig:demo} is a snapshot of our online demo. The system is supported by multiple backend models for test and comparison. Among them, ``Bert-Wiki", corresponding to \te~in this paper, is the best-performing one in our evaluation. After selecting the model(s), users can create their own taxonomy in the ``Labels" column, and input the text snippet. The system will then classify the text with the user-defined taxonomy. Results are presented in two formats: a bar chart and a ranking table. The table on the right provides a clear view of rankings by each model, while the bar chart on the left is useful to compare the scale of the scores from different models for different labels. These scores, ranging from 0 to 1, are probabilities of the label being relevant to the text, which we will explain further in the next section.   

Consider the example in Figure \ref{fig:demo}. The input text is most relevant to \textit{lifestyle}, somewhat relevant to \textit{technology}, and irrelevant to \textit{children}, which aligns with the prediction of our ``Bert-Wiki" model.

\subsection{TE-Wiki}\label{sec: CateTE}
We now describe our best performing model \te. Previous work \cite{YinHaRo19} has demonstrated that an $n$-way classification problem can be converted into $n$ binary entailment problems. Specifically, we can use the text as the premise, and candidate labels as the hypotheses, to generate $n$ statements ``[Text] \textbf{ Entails } [$\textup{Label}_{i}$]" for $i\in [n]$. The motivation is that classification is essentially a special kind of entailment. Suppose we want to classify a document into 3 classes: \textit{politics}, \textit{business}, \textit{sports}. We can ask three binary questions: ``Is it about politics?'', ``Is it about business?'', ``Is it about sports?''. By doing so, the model is no longer constrained to a fixed label set, as we can always ask more questions to handle new labels.

With the above framework, it is straightforward to train a model on an entailment dataset (e.g. MNLI \cite{williams-etal-2018-broad}, FEVER \cite{thorne-etal-2018-fever}, RTE \cite{DaganGlMa05, DBLP:conf/iclr/WangSMHLB19}.) and use it for classification. However, this may not be the optimal choice as topic classification only focuses on high-level concepts, while textual entailment has a much wider scope and involves many other aspects (e.g., see \cite{DRSZ13}).
Therefore, we propose to construct a new dataset from Wikipedia with articles as premises, and categories as hypotheses. Our desired training pair should meet the following two criteria:
\begin{enumerate}[leftmargin=*]
    \vspace{-0.05in}
    \item The hypothesis is consistent with the premise, i.e. the categorization is correct.
    \vspace{-0.05in}
    \item The hypothesis should be abstract and concise to reflect the high-level idea of the premise, rather than focus on certain details.
    \vspace{-0.05in}
\end{enumerate}
Directly using all the categories associated with an article satisfies the first criterion, but fails with the second, as some of them do not represent the article well. For example, the page {\it Bill Gates} is assigned {\it Category:Cornell family}, which is correct about the person but probably not a suitable label for the whole article. To resolve the issue, we instead use higher-level categories on Wikipedia's subcategory graph to yield better hypotheses. 

\begin{algorithm}[t]
\small
\SetAlgoLined
\SetKwInOut{Input}{Input}
\SetKwInOut{Output}{Output}
\Input{Top-level category set $S$, \\
Wikipedia subcategory graph $\mathcal{G}$, \\
max search depth $r=2$\;}
Initialize $d(x, c)=\infty$ for any article $x\in X$ and $c\in S$. $M=\{\}$\;
\For{$c$\textup{ in }$S$}{
    $T$ = DFS$(c, \mathcal{G}, r)$\; 
    \For{$t \textup{ in }T\textup{.nodes}$}{
        \For{$x\textup{ in }t\textup{.articles}$}{
            $d(x,c) = \min\{d(x,c), 1+\textup{depth}(t)\}$\;
        }    
        }
}

\For{$x$\textup{ in }$X$}{
    \If{$\min_{c\in S}d(x,c)<\infty$}{
        $P=\textup{argmin}_{c\in S}d(x,c)$\;
        \For{$c$\textup{ in }$P$}{
            Add $(x, c, 1)$ to $M$\; 
        }
        Sample $c'$ from $S-P$\;
        Add $(x, c', -1)$ to $M$\;
        
    }
}
\Output{$M$}
\caption{Collect training data}
\label{alg: data}

\end{algorithm}

The overview of \te~is illustrated in Figure \ref{fig:model}. Specifically, we start with a set of 700 top-level categories from Wikipedia's overview page\footnote{\url{https://en.wikipedia.org/wiki/Wikipedia:Contents/Categories}} as roots. For each of them, we run a depth-first search (DFS) to find its subcategories. In our experiment, we set the max depth to 2 to ensure the subcategories found are strongly affiliated with the root. We collect all member articles of categories in the DFS tree, including both leaves and internal nodes, and pair them with the root to construct positive examples. In case an article can be reached from multiple root categories, we only pair it with the root(s) that has the smallest tree distance to the article to ensure supervision quality. Then for each article, we randomly choose a different category to construct a negative example. While we have tried more sophisticated negative sampling strategies with the aim to confuse the model, none of them makes a significant improvement. Thus, we keep to this simple version. The final training set $D=\{(x_i, c_i, p_i)_{i=1}^n\}$ consists of 3-tuples such that $x_i$ is a Wikipedia article, $c_i$ is the corresponding high-level category name, and $p_i\in\{+1, -1\}$ is the label. The procedure for constructing the training set is summarized in Algorithm \ref{alg: data}.

We then fine-tune the pre-trained BERT model \cite{DCLT19} with the collected dataset. Given a tuple $(x_i, c_i, p_i)$, the concatenation of $x_i$ and $c_i$ is passed to a BERT encoder, followed by a classification head to predict whether the article $x_i$ belongs to the category $c_i$. During test, (i) for the single-labeled case, we pick the label with the highest predicted probability, (ii) for the multi-labeled case, we pick all labels predicted as positive (i.e. probability $> 0.5$). We do not use any hypothesis template to convert label names into sentences as in \cite{YinHaRo19}, for consistency with training.

%% file: sections/experiment.tex
\section{Evaluation}

We evaluate all the backend models of our system on four classification benchmarks to compare their performance. We also compare them against weakly-supervised and supervised models to quantify how much we can achieve without any domain-specific training data.  

\begin{table}[t]
\small
\centering
\begin{tabular}{lcc}
\hline
Dataset   & \#Classes & \#samples \\ \hline
Yahoo \cite{DBLP:conf/nips/ZhangZL15}     & 10        & 100,000         \\
Situation \cite{MTMWLYFSZYKHSSR19} & 12        & 3525            \\
AG News \cite{DBLP:conf/nips/ZhangZL15}  & 4         & 7,600           \\
DBPedia \cite{DBLP:journals/semweb/LehmannIJJKMHMK15}  & 14        & 70,000          \\ \hline
\end{tabular}
\vspace{-0.05in}
\caption{Dataset Statistics.}
\vspace{-0.15in}
\label{Table: datasets}
\end{table}

\begin{table*}[t]
\small
\centering
\begin{tabular}{clcccc}
\hline
\multicolumn{1}{c}{Supervision Type} & Methods                      & Yahoo         & Situation     & AG News       & DBPedia       \\ \hline
\multirow{6}{*}{Zero-shot}           & Word2Vec \cite{MSCCD13b}                    & 35.7          & 15.6          & 71.1          & 69.7          \\
                                     & ESA \cite{CRRS08}                          & 40.4          & 30.2          & 71.1          & 64.7          \\
                                     & TE-MNLI \cite{YinHaRo19}                     & 37.9          & 15.4          & 68.8          & 55.3          \\
                                     & TE-FEVER \cite{YinHaRo19}                    & 40.1          & 21.0          & 78.0          & 73.0          \\
                                     & TE-RTE \cite{YinHaRo19}                      & 43.8          & 37.2          & 60.5          & 65.9          \\ \cline{2-6} 
                                     & TE-Wiki                  & \textbf{57.3} & \textbf{41.7} & \textbf{79.6} & \textbf{90.2} \\ \hline
\multirow{1}{*}{Weakly-supervised}   
                                     & LOTClass \cite{meng-etal-2020-text}                    & 54.7          & N/A           & 86.4          & 91.1          \\  \hline
Supervised                           & BERT \cite{DCLT19}               & 75.3          & 58.0          & 94.4          & 99.3          \\ \hline
\end{tabular}
\vspace{-0.05in}
\caption{Test results of all methods on four datasets. Compared with Word2Vec and ESA, ESA-WikiCate is overall the best among the three embedding-based methods. TE-WikiCate outperforms all other zero-shot methods across all four datasets, and performs competitively against the weakly-supervised LOTClass.} 
\vspace{-0.15in}
\label{Table: results}
\end{table*}

\subsection{Experiment setup}\label{setup}
\textbf{Datasets: }
We summarize all test datasets in Table \ref{Table: datasets}. For \textit{Yahoo! Answers}, we use the reorganized train/test split by \cite{YinHaRo19}. All datasets are in English.
Among the four, \textit{Situation Typing} is a multi-labeled dataset with imbalanced classes, for which we report the weighted average of per-class F1 score. We refer readers to \cite{YinHaRo19} for the class distribution statistics. The other three are single-labeled and class-balanced, and we report the classification accuracy. 

\noindent\textbf{Models: }
Apart from \te, we run five zero-shot models for open-domain evaluation, as well as a weakly-supervised and a supervised model for close-domain comparison.

\begin{itemize}[leftmargin=*]
    \vspace{-0.05in}
    \item \textbf{Word2Vec} \cite{MSCCD13b}: To measure cosine similarity between the embedding vectors of text and label.
    \vspace{-0.05in}
    \item \textbf{ESA} \cite{CRRS08}:  Same as above, except using embeddings in Wikipedia title space
    \vspace{-0.05in}
    \item \textbf{TE-MNLI}, \textbf{TE-FEVER}, \textbf{TE-RTE} \cite{YinHaRo19}: Textual entailment models by finetuning BERT on MNLI, FEVER, and RTE respectively.\footnote{In experiments, we always use bert-base-uncased.}
    \vspace{-0.2in}
    \item \textbf{LOTClass} \cite{meng-etal-2020-text}: A weakly-supervised method that learns label information from unlabeled, but in-domain training data. 
    \vspace{-0.05in}
    \item \textbf{BERT} \cite{DCLT19}: We finetune a supervised BERT on training data for each dataset. 
    \vspace{-0.2in}
\end{itemize}

\noindent\textbf{Implementation:} We finetune the bert-base-uncased model on the Wikipedia article-category dataset to train \te. We removed 26 categories whose name starts with "List of" from the 700 top-level categories, resulting in 674 categories as hypotheses and 1,367,784 articles as premises. The final training set contains 3,387,028 article-category pairs. We set the max sequence length to be 128 tokens and the training batch size to be 64. The model is optimized with AdamW \cite{DBLP:conf/iclr/LoshchilovH19} with initial learning rate as 5e-5. Since we do not have a development set in the zero-shot setting, we train the model for 1500 steps to prevent overfitting.
For all zero-shot methods, we train once and evaluate the model on all test datasets. For supervised and weakly-supervised methods, we train a different model for each different dataset.

%% file: sections/analysis.tex

\subsection{Result Analysis}
The main results are presented in Table \ref{Table: results}. We observe that \te~performs the best among all zero-shot methods on all four datasets with different labels and from different domains, demonstrating its effectiveness as an open-domain classifier. It also performs closely with the weakly-supervised \textbf{LOTClass} which is trained on in-domain data with known taxonomy, showing that an open-domain zero-shot model can achieve similar accuracy as those domain-specific classifiers. In particular, \te~outperforms \textbf{LOTClass} on \textit{Yahoo}, whose training set contains quite a few ambiguous examples. These examples can have negative impact on self-training. On the other hand, our zero-shot model does not rely on any domain-specific data, making it more robust against imperfect data.

\begin{table*}[t]
\small
\centering
\begin{tabular}{lllll}
\hline
                   & Yahoo      & Situation  & AG News    & DBPedia    \\ \hline
TE-Wiki        & 57.3       & 41.7       & 79.6       & 90.2       \\
- Overlapping categories         & 59.4(+2.1) & 38.8(-2.9) & 79.7(+0.1) & 88.9(-1.3) \\ \hline
Removed training categories & 15         & 2          & 4          & 10          \\
Overlapping test examples (\%)    & 100.0      & 24.0       & 100.0      & 50.0  \\ \hline
\end{tabular}
\caption{Performance before and after removing the overlapping categories, as well as their difference. We also show the number of removed categories, and the percentage of test documents that belong to the overlapping labels. }
\vspace{-0.15in}
\label{Table: overlap}
\end{table*}

It is possible that some of the testing labels also appear in the Wikipedia categories used for training.\footnote{Throughout this subsection, we use the word "category" for the training set and "label" for the testing sets.} To ensure the quality and fairness of our zero-shot evaluation, we remove the overlapping categories from Wikipedia training data, and retrain the \te~model for each test set. Specifically, we normalize labels and categories by their lower-cased, lemmatized names, and perform a token-based matching. We report in Table \ref{Table: overlap} the performance before and after deduplication.

We find that deduplication has little or even positive influence on performance, which shows that \te~does not rely on seeing test labels during training.  
In particular, the performance on \textit{Yahoo} gets improved with deduplication. We suspect that exact match between training and testing labels can lead to overfitting, since the same label may have different meanings under different context.  
Notice that this study is only for justifying our zero-shot evaluation. For real-word applications, excluding overlapping categories is neither necessary nor feasible as users do not know the test labels beforehand in zero-shot scenarios.

\subsection{Early stopping and knowledge transfer}\label{sec: overfit}
To study the convergence of our model, we sample a small dev set of 1000 examples from \textit{Yahoo}'s original validation set. During training, we find that with 25 steps the \te~model already achieves a reasonably good performance on the dev set. Further training for longer steps yields some, but not significant gains. Since the model has only seen $25\times 64=1600$ examples at that point, there is little chance for the model to acquire label specific knowledge with such a small amount of data. Hence, we believe that during the early steps, the model actually learns \textit{``what topic classification is about"}, while the knowledge specific to different labels has already been implicitly stored in the pretrained BERT encoder. The category prediction task takes a minor role in transferring world knowledge. Rather, it teaches the model how to use existing knowledge to make a good inference.

%% file: sections/crowdsourcing.tex
\section{Importance of label names}

\begin{table}[t]
\centering
\small
\begin{tabular}{cccc}
\hline
                   & TE-Wiki  & Human & Supervised \\ \hline
Yahoo              & 58.9 & 64.8  & 77.4       \\
Yahoo-5            & 82.1(+23.2) & 88.1(+23.3)  & 89.4(+12.0)       \\ \hline
AG News            & 79.0 & 80.2  & 94.7       \\
AG News-5          & 86.4(+7.4) & 90.8(+10.6)  & 96.4(+1.7)       \\ \hline
\end{tabular}
\vspace{-0.05in}
\caption{Classification accuracy on crowdsourcing datasets. Yahoo-5 and AG News-5 count only examples for which all five workers choose the same label.}
\vspace{-0.15in}
\label{Table: crowd}
\end{table}

Since zero-shot classifiers understand a label by its name, the quality of label names can be a important performance bottleneck in designing open-domain text classification systems. To study this,
We conduct crowdsourcing surveys on subsets of \textit{Yahoo} and \textit{AG News}. For each dataset, we randomly sample 1,000 documents while preserving class balance. Every document is independently annotated by five workers. 
In the survey question, we only provide the document to be classified and names of candidate labels, without giving workers examples for each class. We consider an example to be correctly classified by humans only if at least three workers choose the gold label. Details about the survey are in Appendix. 

\begin{figure}[t]
    \centering
    \includegraphics[width=0.75\linewidth]{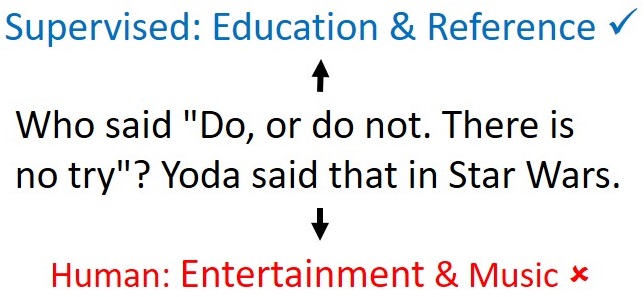}
    \vspace{-0.05in}
    \caption{An example with a bad label name. Annotators are confused by the word ``Reference".}
    \vspace{-0.15in}
    \label{fig:case}
\end{figure}

We summarize the results in Table \ref{Table: crowd}. Row 1\&3 are classification accuracy on the whole crowdsourcing datasets, and row 2\&4 are on subsets of examples where all 5 workers choose the same label. We observe that when including all examples, both \te~and humans perform much worse than the supervised method. The supervised approach has the advantage that it learns data-specific features to resolve ambiguity among different classes. On the other hand, humans only make judgements based on their understanding of the labels and a stand-alone test document, and so does our zero-shot algorithm. Ideally, this task should not be difficult for humans as long as the labels properly describe the text topics. However, in some cases the labels could be ambiguous and confusing.
Figure \ref{fig:case} shows an example of a bad label name leading to a mistake. The word ``Reference" in the correct label actually means ``quoting other people's words". However, it is hard for an ordinary person to understand the meaning without any example as illustration. 4 out of 5 annotators instead chose ``Entertainment \& Music" due to the movie ``Star Wars". By contrast, the supervised model has no difficulty in making the correct decision because it has seen plenty of quotation examples during training and can easily capture the useful pattern like ``Who said XXX". The main reason for humans' confusion here is that the label name does not directly reflect the semantics of the text. A better description of the class should be provided for classification without examples.

We also calculate the accuracy on examples where all 5 workers agree, as in row 2\&4 in Table \ref{Table: crowd}. We believe the high inter-annotator agreement here indicates a better alignment between the semantics of text and label. We find a significant improvement of human performance on these less ambiguous cases. The same happens to our zero-shot model, but the supervised method benefits much less. Consequently, the performance gap between humans and the supervised model is also getting closer, which demonstrates that ambiguous labels have a strongly negative impact on classification. Therefore, we believe picking good labels is crucial for open-domain topic classification.

%% file: sections/conclusion.tex
\section{Conclusion}
We introduce a system for open-domain topic classification. The system allows users to define customized taxonomy and classify text with respect to that taxonomy at real time, without changing the underlying model. To build a powerful model, we propose to utilize Wikipedia articles and categories and adopt an entailment framework for zero-shot learning. The resulting \te~ outperforms all existing zero-shot baselines in open-domain evaluations. Finally, we demonstrate the importance of choosing proper label names in open-domain topic classification through a crowdsourcing study.

%% file: sections/ack.tex
\section*{Acknowledgements}

This work was supported by Contracts FA8750-19-2-1004 and FA8750-19-2-0201 with the US Defense Advanced Research Projects Agency (DARPA). Approved for Public Release, Distribution Unlimited. The views expressed are those of the authors and do not reflect the official policy or position of the Department of Defense or the U.S. Government.

%% file: sections/appendix.tex
\appendix
\section{Crowdsourcing Setup}\label{appendix: crowd}
We conduct crowdsourcing annotations for 1000 documents sampled from the \textit{Yahoo! Answers} dataset and another 1000 from the \textit{AG News} on Amazon Mechanical Turk (AMTurk). Both crowdsourcing subsets preserve the class-balance as in the original datasets. We avoid using long documents so that each document contains no more than 512 characters. The 1000 samples are split into 40 assignments, each containing 25 examples. We request 5 AMTurk workers for multiple-choice questions on each assignment. In order to ensure the response quality, we use anchor examples and gold annotations from the original datasets to filter out low-quality answers. Specifically, in each assignment we insert two anchor examples that we believe are easy enough for workers to choose the correct answer as long as they pay attention. We reject a submission if a worker's classification accuracy against gold annotations is below 30\%, or both anchor examples are wrongly classified. With a small initial pilot, we estimate the average working time for labeling 25 examples to be 22 minutes, and we set the pay rate to be \$1.5 per assignment for each valid submission. The overall cost is \$300 for 200 valid submissions for each dataset.